\pdfoutput=1
\documentclass[11pt]{article}

\usepackage{acl}

\usepackage{times}
\usepackage{latexsym}
\PassOptionsToPackage{breaklinks}{hyperref}

\usepackage[T1]{fontenc}
\usepackage[utf8]{inputenc}

\usepackage{microtype}

\usepackage{hyperref}
\definecolor{darkblue}{rgb}{0, 0, 0.5}
\hypersetup{colorlinks=true, citecolor=darkblue, linkcolor=darkblue, urlcolor=darkblue}

\usepackage{xstring}

\usepackage{algorithm}
\usepackage{algorithmic}
\usepackage{textcomp}
\usepackage{color}
\usepackage{xcolor}
\usepackage{array}
\usepackage{pifont}
\usepackage{tabularx}
\usepackage{adjustbox}
\usepackage{booktabs,dcolumn}
\usepackage{makecell}
\usepackage{multirow}
\usepackage{multicol}
\usepackage{tablefootnote}
\usepackage{threeparttable}
\usepackage{float}
\usepackage{subfigure}
\usepackage{paralist}
\usepackage{comment}
\usepackage{enumerate}
\usepackage{xspace}
\usepackage{mathtools}
\usepackage{amsmath,amsthm,amsfonts,amssymb,bm,stmaryrd,mathrsfs}
\usepackage{url}
\usepackage{mfirstuc,textcase}
\usepackage{fmtcount}
\usepackage{flushend}
\usepackage{thmtools}
\usepackage{colortbl}
\definecolor{mygray}{gray}{.9}
\definecolor{up}{RGB}{238,44,44}
\definecolor{drop}{RGB}{50,205,50}
\usepackage{balance}

\usepackage{CJKutf8}

\usepackage{ulem}

\newcommand{\eg}{\hbox{\emph{e.g.}}\xspace}

\newcommand{\ie}{\hbox{\emph{i.e.}}\xspace}
\newcommand{\wrt}{\hbox{\emph{w.r.t.}}\xspace}

\newcommand{\tabincell}[2]{\begin{tabular}{@{}#1@{}}#2\end{tabular}}

\newtheorem{definition}{Definition} % [section]

\usepackage[most]{tcolorbox}

\newcommand{\classA}{path generation}
\newcommand{\classB}{answer calibration}
\newcommand{\task}{multi-step reasoning}

\title{Towards A Unified View of Answer Calibration for Multi-Step Reasoning}
\author{
  Shumin Deng$^{\clubsuit}$, Ningyu Zhang$^{\heartsuit}$\thanks{$\quad$ Corresponding Author.}~, Nay Oo$^{\clubsuit}$, Bryan Hooi$^{\clubsuit*}$ \\
  $^\clubsuit$National University of Singapore, NUS-NCS Joint Lab ~ $^\heartsuit$Zhejiang University \\
  \texttt{ 
    \{shumin,nay.oo,dcsbhk\}@nus.edu.sg, zhangningyu@zju.edu.cn
  } \\
}

\begin{document}
\maketitle
\normalem

\begin{abstract}
Large Language Models (LLMs) employing Chain-of-Thought (CoT) prompting have broadened the scope for improving multi-step reasoning capabilities. 
We generally divide multi-step reasoning into two phases: \emph{path generation} to generate the reasoning path(s); and \emph{answer calibration} post-processing the reasoning path(s) to obtain a final answer. However, the existing literature lacks systematic analysis on different answer calibration approaches.  
In this paper, we summarize the taxonomy of recent answer calibration techniques and break them down into step-level and path-level strategies. 
We then conduct a thorough evaluation on these strategies from a unified view, systematically scrutinizing step-level and path-level answer calibration across multiple paths. Experimental results reveal that integrating the dominance of both strategies tends to derive optimal outcomes. 
Our study holds the potential to illuminate key insights for optimizing multi-step reasoning with answer calibration. 
\end{abstract}
% summarize the taxonomy of, taxonomize
% yield
% summarize the taxonomy of
% Generally, multi-step reasoning comprises path generation and answer calibration, thereinto, answer calibration is promising while systematic analysis on its effectiveness are still underexplored. 
% Usually, answer calibration strategies such as step-level or path-level calibration play a vital role in multi-step reasoning. 
% While effective, there remains a significant gap in our understanding of the key factors that drive their success. 
% Multi-step reasoning primarily involves two parts: path generation and answer calibration. 
% In this paper, we delve deeper into answer calibration strategies, categorizing them into calibration, applicable to single or multiple paths.
% and present a unified view which establishes associations between step-level and path-level strategies. 

\iffalse
\bibliography{custom}
\bibliographystyle{acl_natbib}
\fi
% !TEX root = ./arr2024.tex

\section{Introduction}
\label{sec:intro}

\begin{figure}[!t] % !htbp
  \centering
  % \vspace{-3mm}
  \includegraphics[width=0.76\linewidth]{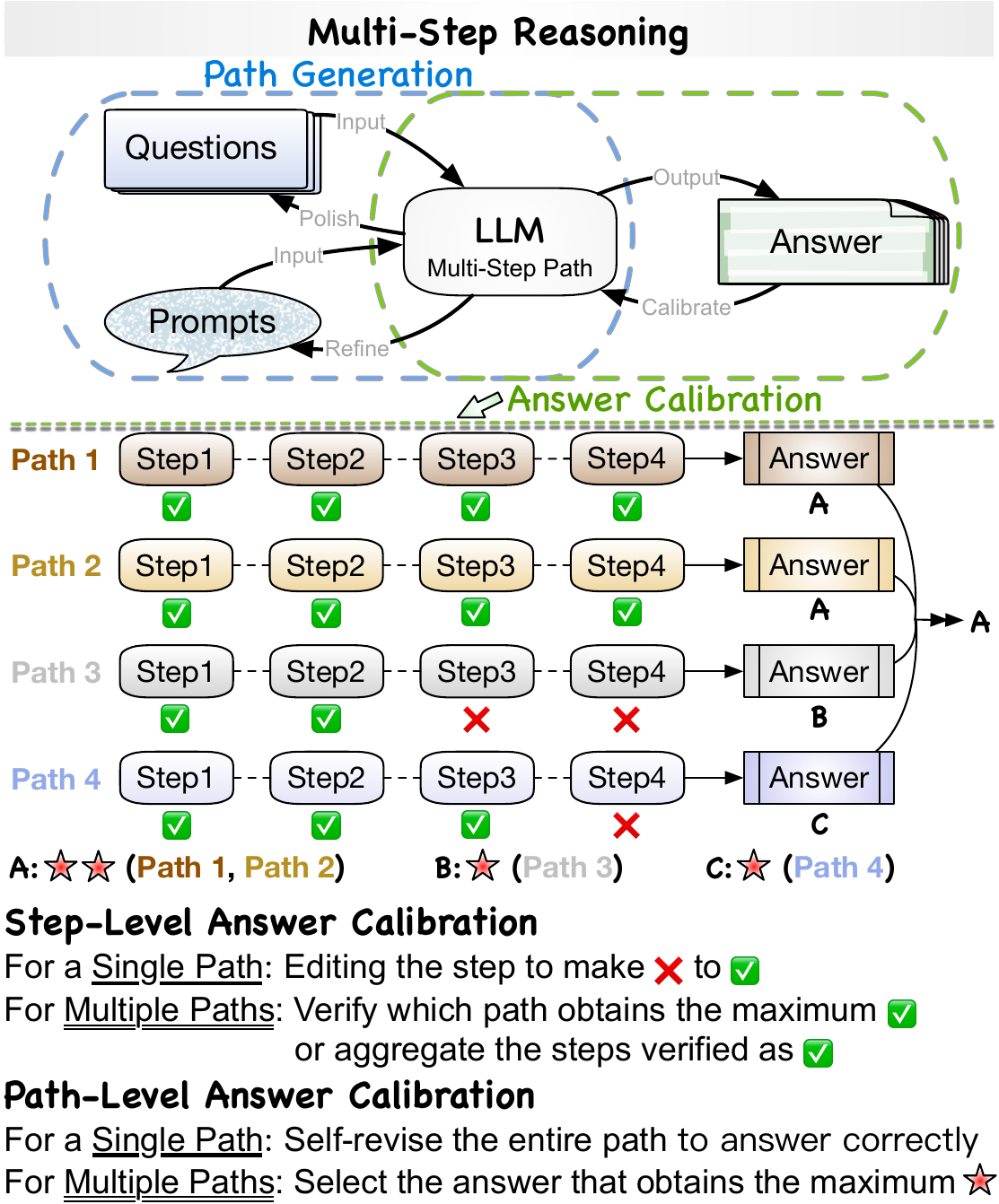} 
  % 0.76, 0.81; 0.85, 0.9
  % \vspace{-7mm}
  \vspace{-3mm}
  \caption{
  Illustration of {\classB} for {\task} with LLM. 
  The methods of step/path-level {\classB} for \emph{multiple paths} can employ {\classB} on \emph{a single path} first. 
  (Terminology clarification of {\classB} and model calibration is elaborated in Appendix~\ref{sec:appendix_clarification_calibration}.)
  \label{fig:intro}
  }
  \vspace{-6mm}
  % \vspace{-2mm}
\end{figure}

Chain-of-Thought (CoT) prompting \cite{NeurIPS2022_CoT} has significantly improved {\task} capabilities of Large Language Models (LLMs) \cite{arXiv2023_LLMSurvey,ACL2023_PromptReasoningSurvey}. 
% Current {\task} methods mainly focus on \emph{generating reasoning process} \cite{ICLR2023_ComplexCoT,NeurIPS2023_ToT} and \emph{calibrating answers} \cite{ICLR2023_Self-Consistency,ACL2023_Verify-and-Edit,EMNLP2023-Findings_Self-Verification} as shown in Figure~\ref{fig:intro}. 
As seen from Figure~\ref{fig:intro}, the process of {\task} generally contains two primary modules: \emph{reasoning} \emph{\classA} which generates one or multiple reasoning paths \cite{ICLR2023_ComplexCoT,NeurIPS2023_ToT}; and \emph{\classB} which post-processes the reasoning path(s) to calibrate the initial output \cite{ICLR2023_Self-Consistency,ACL2023_Verify-and-Edit}. 
% derive the final prediction, rectify  flaws in 
% Given the current lack of a systematic evaluation of {\classB} strategies, this paper seeks to address this gap. 
% Since there still lacks a systematic evaluation of {\classB} strategies, this paper aims to fill the gap. 
% this paper primarily encompasses evaluation on {\classB} strategies. 

In practice, {\classB} is pluggable and can be integrated into {\classA} models. % incorporated
%Our overall framework divides {\classB} into step and path levels, applicable to single or multiple paths, as illustrated in Figure~\ref{fig:intro}. % categorize
The {\classB} framework can be divided into step and path levels, applicable to single or multiple paths, as illustrated in Figure~\ref{fig:intro}. % categorize
For \emph{step-level} {\classB} on {\uline{a single path}}, the model rectifies errors in intermediate-step answers of a generated path \cite{ACL2023_Verify-and-Edit}.  
For \emph{step-level} {\classB} on {\uuline{multiple paths}}, the model verifies each intermediate-step answer \cite{EMNLP2023-Findings_Self-Verification} or aggregates the correct step answers \cite{arXiv2023_GraphVerifier} from multiple paths.   
For \emph{path-level} {\classB} on {\uline{a single path}}, the model revises the entire rationale to obtain the correct answer \cite{EMNLP2023_KALMV}.  
For \emph{path-level} {\classB} on {\uuline{multiple paths}}, the model produces a result indicating the consensus of all candidate paths \cite{ICLR2023_Self-Consistency,EMNLP2023_MCR}. 
% gives the result with overall consideration of all candidate path answers. 
% considering all candidate path answers. or with overall consideration of all candidate path answers. 
As {\classB} can identify and rectify errors in the reasoning path, or even holistically utilize multiple candidate paths, it plays a vital role in {\task} to ensure a precise, consistent and reliable reasoning process \cite{TACL2024_Survey_Self-Correction}.

However, we argue that the crucial factors driving the success of {\classB} strategies remain obscure, with a comprehensive systematic analysis still underexplored. 
To bridge the gap, our study investigates: 
% in this paper, we delve into the following questions: 
(1) The specific conditions where {\classB} notably boosts {\task} performance;  
(2) The strengths and weaknesses of step-level versus path-level {\classB}, and the pathway to attaining optimal performance; 
(3) The robustness and generalizability of {\classB} strategies. 
% enigmatic
% advantages and disadvantages
% (1) What are the underlying connections between these strategies?
% (2) Are there common design elements that play a pivotal role in their effectiveness, and if so, what are they?
% (3) Is it possible to transfer the successful components of each strategy to enhance the efficacy of others?

% providing a systematic analysis of step-level and path-level {\classB} across multiple paths. 
To address these questions, we dissect cutting-edge {\classB} techniques for {\task} with LLMs, and introduce a unified framework that elucidates  step-level and path-level strategies.
% In this paper, we take the first step to provide a unified view of {\classB} for reasoning with LLMs. 
We define two thresholds to respectively signify the step-level and path-level dominance in the unified framework.
We then undertake a comprehensive evaluation of {\classB} strategies, \wrt accuracy, faithfulness, informativeness, consistency, and perplexity over steps or paths. 
% relevance, concordance and consistency over steps or paths. 
% from various perspectives on several {\task} tasks
% a series of 
% strategies 
Through rigorous experiments on five representative {\task} tasks involving arithmetic \cite{EACL2024-StuWS_Survey-MathematicalReasoning} and commonsense, we find that:  
\textbf{(1)} employing {\classB} can enhance accuracy, with the improvement being more noticeable in zero-shot scenarios ($\S$\ref{sec:exp_strategy}) and less significant on stronger backbone models ($\S$\ref{sec:exp_backbone}); 
\textbf{(2)} The optimal performance of the unified {\classB} strategy typically achieved by synthesizing step-level and path level dominance ($\S$\ref{sec:exp_strategy_joint}); 
% , but dominant step-level {\classB} can cause the best results to deviate from this range ($\S$\ref{sec:exp_strategy_joint}); 
\textbf{(3)} path-level {\classB} is more beneficial in improving accuracy, and step-level {\classB} is more effective for mitigating low-quality prompting ($\S$\ref{sec:exp_prompting});   
\textbf{(4)} {\classB} can improve consistency on arithmetic tasks but weakens faithfulness, informativeness and perplexity on both arithmetic and commonsense tasks ($\S$\ref{sec:exp_tasks}).

% such as without coherence and relevance
% and the outperformance will be more prominent in zero-shot scenarios 
% within steps and I/O
% Our objective is to shed light on the key insights that can guide the optimization of multi-step reasoning through effective {\classB}. We hope that our findings will contribute to the ongoing efforts to enhance the capabilities of LLMs and their application in multi-step reasoning.

\iffalse
\bibliography{custom}
\bibliographystyle{acl_natbib}
\fi
% !TEX root = ./arr2024.tex

\section{Related Work}
\label{sec:related_work}

\textbf{Reasoning Path Generation.} 
Previous methods for reasoning {\classA} mostly focus on two aspects to improve reasoning process, including refining input query or prompts (\emph{input refinement}) and polishing the reasoning path (\emph{rationale polish}). 

As for \emph{input refinement}, 
Zero-shot CoT \cite{NeurIPS2022_ZeroShotCoT} and Few-shot CoT \cite{NeurIPS2022_CoT} are classic methods to elicit {\task} ability of LLMs, with ``Let's think step by step'' prompts. 
% or prompting with demonstrations
To decouple planning and execution, \citet{ACL2023_Plan-and-Solve,EACL2024_Plan-Act} devise a plan by prompting and divide and conquer multi-step tasks. 
% propose plan-and-solve prompting by devising a plan to divide and conquer multi-step tasks. 
To enrich prompts, \citet{ACL2024_CoK-Prompting} leverage structure triples as evidence, \citet{NAACL2024_Role-Play-Prompting} design role-play prompting, and \citet{arXiv2023_Re-Reading} employ re-reading instructions. 
% \citet{ICLR2023_ComplexCoT} demonstrated that prompts with higher reasoning complexity can substantially improve LLM performance. 
Besides, LLM performance can also be affected by prompt complexity \cite{ICLR2023_ComplexCoT} and formats, such as program \cite{ICML2023_PAL,TMLR2023_PoT,ICML2024_AoT,arXiv2023_DesignOfCoT,arXiv2023_SelfzCoT,AAAI2024_AnalyzePoT} and table \cite{ACL2023-Findings_Tab-CoT}. 
Further, \citet{EMNLP2023-Findings_Self-promptedCoT,arXiv2023_PromptSpaceOpt,COLING2024_MinT} propose to adaptively utilize prompts. 
Apart from refining prompts, \citet{EMNLP2023-Findings_Self-Polish} progressively refine the given questions, \citet{ACL2024-Findings_Meta-Reasoning} convert semantically-wrapped questions to meta-questions, and \citet{ACL2023-Findings_DyProgramPrompting} augment training data with program annotations. 
%  via semantic resolution

In terms of \emph{rationale polish}, recent work mainly focus on step-aware training \cite{arXiv2023_InteractiveNLP} and path-level optimization. 
For step-aware training, 
\citet{ACL2023_Step-by-StepPlanning} introduce step-by-step planning and \citet{ACL2023-Findings_RoT} recursively tackle intermediate steps;  
\citet{arXiv2023_Resprompt} reconstruct the reasoning rationale within prompts by residual connections;  
\citet{EACL2024_REFINER} iteratively provide feedback on step answers;  
\citet{NeurIPS2023_Self-Notes} leverage self-notes as intermediate steps and working memory; 
% and enabled take notes deviating from input context. 
\citet{ACL2023_DIVERSE,NeurIPS2023_DeductiveVerfication,ICLR2024_VerifyStepbyStep} propose to verify on intermediate step answers;  
% proposed a diverse verifier with voting on reasoning steps. 
\citet{ICLR2024_CoK,arXiv2023_KDCoT} process step-aware verification by knowledge base retrieval. 
% leveraged factual statements for, knowledge-intensive CoT
For path-level optimization, \citet{EMNLP2023_MoT} enable LLMs to self-improve via pre-thinking and recalling relevant reasoning paths as memory;  
\citet{ICLR2024_MathCoder,ICLR2024_MAmmoTH} leverage hybrid rationales in formats of natural language and program.   
Some work also generate deliberate rationales beyond CoT, such as Tree-of-Thought \cite{NeurIPS2023_ToT,arXiv2023_ToT_2}, Graph-of-Thought \cite{arXiv2023_GoT,arXiv2023_GoT_2} and Hypergraph-of-Thought (HoT) \cite{arXiv2023_HoT}. 

\noindent
\textbf{Answer Calibration.} 
Given generated reasoning path(s), {\classB} methods \emph{post-process} the path(s) to calibrate the answer, involving step- or path-level calibration on one or multiple path(s). 

\emph{Step-level {\classB}}.  
\citet{arXiv2023_RCOT,arXiv2023_GraphVerifier} propose to rectify factual inconsistency and reasoning logic between intermediate steps. 
% \citet{arXiv2023_RCOT} detect and rectify step-wise factual inconsistency in the reasoning path. 
% \citet{arXiv2023_GraphVerifier} propose to verify reasoning logic between intermediate steps. 
% % a reasoning graph verifier 
% % between original and reversingly-reconstructed solutions.  
\citet{ICLR2024_Self-Check,AAAI2024_Verify-then-Rectify} check the correctness of each intermediate step. 
\citet{ACL2023_Verify-and-Edit} post-edit {\task} paths with external knowledge. 
% propose a verify-and-edit framework to
% to increase prediction factuality 
\citet{ICLR2023_ReAct,EMNLP2023_RAP,NeurIPS2023_Reflexion,ICLR2024_Retroformer,arXiv2023_FireAct,arXiv2023_ReSTReAct} draw up a plan and act step by step with LLMs as agents \cite{FCS2024_Survey-Agent,arXiv2023_AgentSurvey_2}, encouraging interaction with the environment to provide feedback. 
\citet{EMNLP2023-Findings_Self-Verification,ACL2024-Findings_Forward-BackwardReasoning} unleash the self-verification ability of LLMs, by forward reasoning and backward verification on intermediate step answers. 
\citet{ICLR2024_CSV} propose code-based self-verification on reasoning steps. 
% explicit

\emph{Path-level {\classB}}.  
\citet{NeurIPS2022_STaR} present a self-taught reasoner to iteratively generate rationales. 
% iteratively leverage a few examples to generate many reasoning paths and finetuned on all of them. 
% present a self-taught reasoner to iteratively generate rationales. 
\citet{arXiv2023_ProHint-Prompting} progressively use the generated answers as hints to make double-check. 
\citet{ECML-PKDD2023_GPT-LODS} enrich generated reasoning paths with hundreds of RDF KGs for fact checking. 
\citet{EMNLP2023_KALMV} iteratively rectify errors in knowledge retrieval and answer generation for knowledge-augmented LMs. 
To cultivate the reasoning ability of smaller LMs, \citet{ACL2023_ReasoningTeachers,ACL2023_SCOTT,EMNLP2023_TailoredLearning} propose to fine-tune CoT for knowledge distillation. 
\citet{EMNLP2023_Self-Improve} demonstrate that LLMs can self-improve with high-confidence rationale-augmented answers. 
\citet{EMNLP2023_MCR} prompt LLMs to meta-reason over multiple paths. 
% introduce multi-chain reasoning by 
\citet{arXiv2023_CoH,ICLR2024_Self-Refine} leverage feedback to improve model initial outputs. 
% \citet{arXiv2023_CoH} leverage all available feedback data to improve model performance. 
% \citet{ICLR2024_Self-Refine} introduced a Self-Refine framework to improve initial outputs from LLMs through iterative feedback and refinement.
\citet{ACL2023-Findings_Self-AdaptivePrompting} adaptively select in-context demonstrations from previous outputs to re-generate answers. 
\citet{ICLR2023_Self-Consistency} leverage self-consistency decoding strategy to majority vote on multiple path answers. 
% \citet{EMNLP2023_MoT}
\citet{EMNLP2023_Sample-StepbyStep} propose adaptive-consistency to reduce sample budget.

% \input{preliminary}

\iffalse
\bibliography{custom}
\bibliographystyle{acl_natbib}
\fi
% !TEX root = ./arr2024.tex

\section{Comprehensive Analysis of Answer Calibration}
\label{sec:method}

% \capitalisewords
% \textbf{\makefirstuc{\classB}}
\subsection{Formulation of Answer Calibration} 

Given a question denoted as $\mathcal{Q}$ and its associated prompt $P$, we leverage the LLM to generate the result $\mathcal{R}$. $\mathcal{R}$ can either encompass a single reasoning path $\mathcal{P}$ with an initial answer $\mathcal{A}$ or multiple reasoning paths $\mathbb{P} = \{ \mathcal{P}_i \}_{i \in [1, N]}$ with a corresponding answer set $\mathbb{A} = \{ \mathcal{A}_i \}_{i \in [1, N]}$. The total number of paths in $\mathbb{P}$ is $N$. 
In this paper, we analyze under the assumption that each reasoning path comprises a maximum of $M$ steps. Paths exceeding $M$ steps are truncated, and those with fewer steps are padded. The intermediate step answers for each reasoning path $\mathcal{P}_{(i)}$ are represented as ${\{a_j\}}^{(i)}_{j \in [1, M]}$. 
% Reasoning paths exceeding this limit are truncated, while those with fewer steps are padded. 
% contains up to 

\paragraph{Step-Level Answer Calibration.} 
Given a single reasoning path $\mathcal{P}$ with an initial final path answer $\mathcal{A}$ and intermediate step answers ${\{a_j\}}_{j \in [1, M]}$, the objective of step-level {\classB} is to rectify any erroneous $a_j$, so that deriving the correct $\mathcal{A}$. 
For multiple reasoning paths $\mathbb{P}$, step-level {\classB} seeks to either select the reasoning path with the maximum correct intermediate step answers or aggregate the verified correct steps to form the most accurate reasoning path, leading to a correct final path answer. 
\emph{Self-verification} \cite{EMNLP2023-Findings_Self-Verification} is an effective approach for step-level {\classB} on multiple reasoning paths. 

\paragraph{Path-Level Answer Calibration.}
Given a single reasoning path $\mathcal{P}$ with an initial final path answer $\mathcal{A}$, the goal of path-level {\classB} is to revise the wrong $\mathcal{A}$. 
For multiple reasoning paths $\mathbb{P} = \{ \mathcal{P}_i \}_{i \in [1, N]}$ with corresponding answers $\mathbb{A} = \{ \mathcal{A}_i \}_{i \in [1, N]}$, path-level {\classB} is designed to select the reasoning path from $\mathbb{P}$ with the most consistent answer in $\mathbb{A}$. 
\emph{Self-consistency} \cite{ICLR2023_Self-Consistency} is a widely-used efficacious technique for path-level {\classB} on multiple reasoning paths. 
% time-honored and widely-used, classical ("classical" is not really appropriate for work from 2023 (maybe at least 3 years))

% \paragraph{Proposal: Jointly step-level and path-level {\classB} on multiple paths.} 
\subsection{Unified View of Answer Calibration} 
\label{sec:unfied-view}
Considering the advantages of both step-level and path-level {\classB}, we propose to integrate the two strategies on multiple paths. Given the multiple generated reasoning paths $\mathbb{P} = \{ \mathcal{P}_i \}_{i \in [1, N]}$, we define a unified score $\mathcal{D}_i$ for each $\mathcal{P}_i$ (with the final path answer: $\mathcal{A}_i$ and intermediate step answers: ${\{a_j\}}^{(i)}_{j \in [1, M]}$): 
\begin{equation}
\mathcal{D}_i ~~= \underbrace{ \alpha \frac{n_i}{N} }_{path-level} + ~~~
\underbrace{( 1 - \alpha) \frac{m_i}{M} }_{step-level}
\label{eq:measure_score}
\end{equation}
% \alpha F_i + (1 - \alpha) S_i = 
where $n_i \in [1, N]$ is the frequency of $\mathcal{A}_i$ existing in $\mathbb{A}$, $m_i \in [0, M]$ is the number of correct intermediate steps in $\mathcal{P}_i$, and $\alpha$ is a hyper-parameter. 
\emph{The final answer is $\mathcal{A}_{i^*}$ satisfying $i^* = \mathop{\arg\max}\limits_{i \in [1, N]}(\mathcal{D}_i)$.}
% $i^* = \argmax_{i \in [1, N]} (\mathcal{D}_i)$
% \emph{The final answer is $\mathcal{A}_x$ of the reasoning path $\mathcal{P}_x$ which achieves the maximum score $D_x$ among all path scores, satisfying $x = \argmax_i(\mathcal{D}_i)$.}

% Given the formulation of {\classB}, we then quantify the dominance of step-level and path-level calibration in the unified framework. 
To better analyze the effects of varying $\alpha$ in the unified framework, we then define particular choices for $\alpha$ which we call \emph{step and path level dominant {\classB}}.  
% in terms of {\task} performance. 
% mathematical, establish a unified framework to 

% (1) 
\begin{definition}
\textbf{Step-Level Dominant Answer Calibration}: 
% For this choice, the step-level score in the unified $\mathcal{D}_i$ is always prioritized, 
This choice refers to the level of $\alpha$ at which the step-level score is used as the dominant criterion, 
with the path-level score given much smaller weight and only serving to break ties when necessary. Specifically, larger $m_i$ always results in larger $\mathcal{D}_i$, no matter how small $n_i$ is. We denote this as: 
$\forall n_j, n_k \in [1, N] ~ \text{and} ~ m_j, m_k \in [0, M], \text{where} ~ n_j < n_k ~ \text{and} ~ m_j > m_k$, the scores $D_j$ and $D_k$ should satisfy 
$$
\alpha \frac{n_j}{N} + (1 - \alpha) \frac{m_j}{M} > \alpha \frac{n_k}{N} + (1 - \alpha) \frac{m_k}{M} 
\label{eq:measure_score_sv_sc} 
$$
\end{definition}
Thus we can obtain 
\begin{equation}
\alpha <  \frac{1} { \frac{ M(n_k-n_j) }{ N(m_j-m_k)} + 1 } 
\label{eq:measure_score_sv_sc_alpha}
\end{equation}
% \alpha < \frac{ N(m_j-m_k) }{ M(n_k-n_j) + N(m_j-m_k) }
% considered first

If Eq~\eqref{eq:measure_score_sv_sc_alpha} is constant, we can infer that 
\begin{equation}
\alpha < \min \left( \frac{1} { \frac{ M(n_k-n_j) }{ N(m_j-m_k)} + 1 } \right)
 = \frac{1} { \frac{ M \max(n_k-n_j) }{ N \min(m_j-m_k)} + 1 }
\label{eq:measure_score_sv_sc_alpha_min}
\end{equation}
As $1 \leq n_j < n_k$, $n_j + n_k \leq N$, and $0 \leq m_k < m_j$, we can deduce that $\min(m_j - m_k) = 1$, $\max(n_k - n_j) = N - 2$. % = (N - 1) - 1
From the above, we deduce: 
\begin{equation}
\alpha < \frac{1}{ \frac{M(N-2)}{N} + 1 }
\label{eq:measure_score_sv_sc_alpha_min_final}
\end{equation} 
% $\min(m_j - m_k) = 1$, $\max(m_j - m_k) = M$; $\min(n_k - n_j) = 1$, $\max(n_k - n_j) = N - 2$
% Thus $\max \left( \frac{n_k-n_j}{m_j-m_k} \right) = \frac{\max(n_k - n_j)}{\min(m_j - m_k} = N - 2$. 

% Conversely, 
% (2) 
% This choice gives priority to the path-level score in the unified $\mathcal{D}_i$,
% 
\begin{definition}
\textbf{Path-Level Dominant Answer Calibration}: 
For this choice, $\mathcal{D}_i$ gives priority to the path-level score, with the step-level score given much smaller weight and only serving to break ties when necessary. Concretely, larger $n_i$ always conduces larger $\mathcal{D}_i$, no matter how small $m_i$ is. We denote this as: 
$\forall n_j, n_k \in [1, N] ~ \text{and} ~ m_j, m_k \in [0, M], \text{where} ~ n_j > n_k ~ \text{and} ~ m_j < m_k$, the scores $D_j$ and $D_k$ should  satisfy 
$$
\alpha \frac{n_j}{N} + (1 - \alpha) \frac{m_j}{M} > \alpha \frac{n_k}{N} + (1 - \alpha) \frac{m_k}{M} 
$$
\end{definition}
Analogously, we can obtain 
\begin{equation}
\alpha >  \frac{1} { \frac{ M(n_j-n_k) }{ N(m_k-m_j)} + 1 } 
\label{eq:measure_score_sc_sv_alpha}
\end{equation}
If Eq~\eqref{eq:measure_score_sc_sv_alpha} is constant, we can infer that 
\begin{equation}
\alpha > \max \left( \frac{1} { \frac{ M(n_j-n_k) }{ N(m_k-m_j)} + 1 } \right) 
= \frac{1} { \frac{ M \min(n_j-n_k) }{ N \max(m_k-m_j)} + 1 } \
\label{eq:measure_score_sc_sv_alpha_max}
\end{equation}
As $1 \leq n_k < n_j$, and $0 \leq m_j < m_k \leq M$, we deduce that $\min(n_j - n_k) = 1$, $\max(m_k - m_j) = M - 0 = M$. 
From the above, we deduce: 
\begin{equation}
\alpha > \frac{1}{ \frac{1}{N} + 1}
\label{eq:measure_score_sc_sv_alpha_max_final}
\end{equation}

In general, considering \emph{step-level and path-level {\classB} dominance}, we can obtain two thresholds: $\frac{1}{ \frac{M(N-2)}{N} + 1 }$ and $\frac{1}{ \frac{1}{N} + 1}$. 
Note that \textbf{$\bm{\alpha = 0}$ and $\bm{\alpha = 1}$ are respectively equivalent to the self-verification and self-consistency strategies}.

% Calculation of \texttt{ROSCOE} Scores.
\subsection{Evaluation of Answer Calibration} 
\label{sec:ROSCOE}
Calculation of \texttt{ROSCOE} Scores. In addition to the classical evaluation metric: Accuracy, \citet{ICLR2023_ROSCOE} have proposed \textbf{\texttt{ROSCOE}}, a suite of metrics for {\task}, under four perspectives: 
semantic alignment (\texttt{ROSCOE-SA}), 
semantic similarity (\texttt{ROSCOE-SS}), 
logical inference, and (\texttt{ROSCOE-LI}) and 
language coherence (\texttt{ROSCOE-LC}). 
Due to space limits, we select some representative scores from \texttt{ROSCOE} as evaluation metrics in the experiments. 
% We select some scores from \texttt{ROSCOE} to do evaluation in our paper for space limits. Furthermore, these scores also show noticeable difference among various methods.  
% % fine-grained

Given source ground truth rationale ($\bm{s}$) and generated rationale ($\bm{h}$) with multiple steps ($h_i$), we calculate five scores (\emph{All scores satisfy the principle that larger is better}): 

(1) Faithfulness$_{step}$ ($\bm{h} \rightarrow \bm{s}$): 
% To measure if the model misinterpreted the problem statement, or the reasoning path is too vague, irrelevant, or misuses information. 
To assess whether the model misconstrues the problem statement, or if the reasoning path is too nebulous, irrelevant, or improperly employs input information.
\begin{equation}
\textstyle\sum_{i=1}^{N}r\textnormal{-align}(h_i \rightarrow \bm{s}) / N 
\label{eq:faithfulness_step}
\end{equation}
where $N$ is the number of steps and $r\textnormal{-align}$ is used to measure how well $h_i \in \bm{h}$ can be aligned with any one of the steps in the ground truth path $\bm{s}$. 
% $r\textnormal{-align}$ is the alignment vector from steps in $\bm{h}$ to all steps in $\bm{s}$. 
% denotes the reasoning alignment score. 
% To measure alignment from the hypothesis steps to the source sentences, by calculating the mean reasoning alignment score over the steps of reasoning。 

(2) Informativeness$_{path}$ ($\bm{h} \rightarrow \bm{s}$): 
To measure the level of concordance between the generated path and the source, and if the generated reasoning path is well-grounded with respect to the source. 
\begin{equation}
[1 + \cos(\bm{h}, \bm{s})] / 2 
\label{eq:infor_path}
\end{equation} 
where $\cos(\cdot, \cdot)$ is a function for cosine similarity. 
% To quantify the degree of agreement between the hypothesis path and the source

(3) Consistency$_{steps}$ ($h_i \leftrightarrow h_j$): 
To measure logical entailment errors \emph{within} the reasoning steps. 
\vspace{-3mm}
\begin{equation}
1-\textstyle\max_{i=2..N} \textstyle\max_{j<i}p_{\mathrm{contr}}(h_i , h_j) 
\label{eq:consistency_step}
\end{equation} 
where $p_{\mathrm{contr}}$ is used to assess the likelihood of step pairs contradicting each other. $h_i \in \bm{h}$ and $h_j \in \bm{h}$. 

(4) Consistency$_{path}$ ($\bm{h} \leftrightarrow \bm{s}$): 
To evaluate mistakes in logical entailment between the generated reasoning path $\bm{h}$ and source context $\bm{s}$: 
\begin{equation}
1 - \textstyle\max_{i=1..N} \textstyle\max_{j=1..T} ~p_{\mathrm{contr}}(h_i , s_j)
\label{eq:consistency_path}
\end{equation} 
where $p_{\mathrm{contr}}$ is the likelihood of source and generated steps contradicting each other. $s_j \in \bm{s}$; $h_i \in \bm{h}$. 

(5) Perplexity$_{path}$ ($\bm{h}$): 
As an indicator of language coherence, it calculates average perplexity of all tokens in the generated reasoning path steps. 
\begin{equation}
1 / \text{PPL}(\bm{h}) 
\label{eq:perplexity_path}
\end{equation} 
where $\text{PPL}$ denotes the perplexity. 
% The context used to score each token is the previous tokens in the current and all previous steps. Steps are joined with a space character. To keep the range and orientation consistent with the other scores we invert the perplexity. 

\iffalse
\bibliography{custom}
\bibliographystyle{acl_natbib}
\fi
% !TEX root = ./arr2024.tex

\iffalse
\bibliography{custom}
\bibliographystyle{acl_natbib}
\fi
% !TEX root = ./arr2024.tex

\begin{table*}[!t] %[!htbp]
\centering
\small

% \vspace{-3mm}

\resizebox{\textwidth}{!}{%
\begin{tabular}{c | l | l l l l l l}

\toprule

\textbf{Task} & \multicolumn{1}{c|}{\textbf{Method}} & 
 \multicolumn{1}{c}{\textbf{Accuracy} $\uparrow$} & 
% \tabincell{c}{ \textbf{Repetition} \\ (Over Tokens) } & 
% \textbf{Repetition} & % $_{Token}$ & 
\tabincell{c}{ \textbf{Faithfulness} $\uparrow$ \\ (Over Steps) } & 
% \textbf{Faithfulness} & 
\tabincell{c}{ \textbf{Informativeness} $\uparrow$ \\ (Over Path) } &
% \textbf{Info-Path}  & % $_{path}$
\tabincell{c}{ \textbf{Consistency} $\uparrow$ \\ (Within Steps) } & 
% \textbf{Self-Consistency}
\tabincell{c}{ \textbf{Consistency} $\uparrow$ \\ (Within I/O) } & % \textbf{Source-Consistency}
\tabincell{c}{ \textbf{Perplexity} $\uparrow$ \\ (Over Path) }
% \textbf{Perplexity} 
\\

\midrule

\multicolumn{1}{c|}{ \multirow{7}{*}{ \textbf{GSM8K} } } 
% \multicolumn{1}{c|}{ \multirow{7}{*}{ \textbf{Arithmetic} } } 
% \multicolumn{1}{c|}{ \multirow{7}{*}{ \tabincell{c}{ \textbf{Arithmetic} \\ (GSM8K) \\} } } 
% & FT SOTA 	& 57$^\alpha$ & \quad/ & \quad/ & \quad/ & \quad/ & \quad/ \\
% \cmidrule{2-8}
% \;\,

& CoT & 80.21   
& 88.73
& 96.38 
& \textbf{97.94} 
& 96.94 
& 9.14 
\\

& CoT + SV 	& 82.34$_{\color{up}{(+2.13)}}$ 
& 86.22$_{\color{drop}{(-2.51)}}$
& 95.19$_{\color{drop}{(-1.19)}}$
& 96.78$_{\color{drop}{(-1.16)}}$
& 93.46$_{\color{drop}{(-3.48)}}$
& \textbf{14.90}$_{\color{up}{(+5.76)}}$
\\

& CoT + SC 	& \textbf{87.11}$_{\color{up}{(+6.90)}}$ 
& \textbf{88.83}$_{\color{up}{(+0.10)}}$
& \textbf{96.40}$_{\color{up}{(+0.02) \sim}}$
& 97.90$_{\color{drop}{(-0.04) \sim}}$
& \textbf{97.44}$_{\color{up}{(+0.50)}}$
& 8.90$_{\color{drop}{(-0.24)}}$ 
\\

% % \cdashline{3-8}[1pt/1pt]
\cmidrule{2-8}

& ZS CoT 		& 62.85     
& 86.58 
& 95.61  
& \underline{97.30} 
& 93.07 
& \underline{15.67} 
\\

& ZS CoT + SV 	& 67.70$_{\color{up}{(+4.85)}}$ 
& 86.24$_{\color{drop}{(-0.34)}}$
& 95.19$_{\color{drop}{(-0.42)}}$
& 96.78$_{\color{drop}{(-0.52)}}$
& 93.44$_{\color{up}{(+0.37)}}$
& 14.90$_{\color{drop}{(-0.77)}}$
\\

& ZS CoT + SC 	& \underline{71.42}$_{\color{up}{(+8.57)}}$ 
& \underline{86.70}$_{\color{up}{(+0.12)}}$
& \underline{95.67}$_{\color{up}{(+0.06) \sim}}$
& 97.19$_{\color{drop}{(-0.11)}}$
& \underline{94.57}$_{\color{up}{(+1.50)}}$
& 14.95$_{\color{drop}{(-0.72)}}$
\\

% & P-S       & 61.26     
% & 84.89 
% & 94.57  
% & 97.63 
% & 92.26 
% & 13.84 
% \\

% & P-S + SV   & -$_{\color{drop}{(-0.01)}}$ 
% & -$_{\color{drop}{(-0.01)}}$
% & -$_{\color{drop}{(-0.01)}}$
% & -$_{\color{drop}{(-0.01)}}$
% & -$_{\color{drop}{(-0.01)}}$
% & -$_{\color{drop}{(-0.01)}}$
% \\

% & P-S + SC   & -$_{\color{drop}{(-0.01)}}$ 
% & -$_{\color{drop}{(-0.01)}}$
% & -$_{\color{drop}{(-0.01)}}$
% & -$_{\color{drop}{(-0.01)}}$
% & -$_{\color{drop}{(-0.01)}}$
% & -$_{\color{drop}{(-0.01)}}$
% \\

\midrule

\multicolumn{1}{c|}{ \multirow{7}{*}{ \textbf{SVAMP} } } 
% & FT SOTA 	& 57.4$^\beta$ & \quad/ & \quad/ & \quad/ & \quad/ & \quad/ \\
% \cmidrule{2-8}

& CoT       & 78.20 
& \textbf{87.73}
& \textbf{95.74} 
& 30.57
& 9.82 
& \textbf{6.65} 
\\

& CoT + SV 	& \textbf{85.80}$_{\color{up}{(+7.60)}}$ 
& 87.26$_{\color{drop}{(-0.47)}}$
& 95.00$_{\color{drop}{(-0.74)}}$
& 33.39$_{\color{up}{(+2.82)}}$
& \textbf{10.41}$_{\color{up}{(+0.59)}}$
& 6.23$_{\color{drop}{(-0.42)}}$
\\

& CoT + SC 	& 84.40$_{\color{up}{(+6.20)}}$ 
& 87.60$_{\color{drop}{(-0.13)}}$
& 95.71$_{\color{drop}{(-0.03)}}$
& \textbf{33.51}$_{\color{up}{(+2.94)}}$
& 9.92$_{\color{up}{(+0.10)}}$
& 6.22$_{\color{drop}{(-0.43)}}$
\\

\cmidrule{2-8}

% & ZSL CoT 	& 72.20 
% & 87.39
% & 95.77 
% & 32.53 
% & 19.01 
% & 11.90 
% \\

& ZS CoT        & 72.80     
& \underline{87.46} 
& 95.77  
& 31.71 
& 18.39 
& \underline{11.93} 
\\

& ZS CoT + SV   & 81.20$_{\color{up}{(+8.40)}}$ 
& 86.92$_{\color{drop}{(-0.54)}}$
& 95.05$_{\color{drop}{(-0.72)}}$
& \underline{35.27}$_{\color{up}{(+3.56)}}$
& \underline{20.24}$_{\color{up}{(+1.85)}}$
& 11.44$_{\color{drop}{(-0.49)}}$
\\

& ZS CoT + SC   & \underline{82.00}$_{\color{up}{(+9.20)}}$ 
& 87.40$_{\color{drop}{(-0.06)}}$
& \underline{95.81}$_{\color{up}{(+0.04) \sim}}$
& 34.73$_{\color{up}{(+3.02)}}$
& 19.67$_{\color{up}{(+1.28)}}$
& 11.68$_{\color{drop}{(-0.25)}}$
\\

% & P-S 		& \underline{68.00}     
% & 81.44 
% & \underline{94.78}  
% & 22.57 
% & 14.44 
% & \underline{12.65} 
% \\

% & P-S + SV 	& 61.30$_{\color{drop}{(-6.70)}}$ 
% & 85.18$_{\color{up}{(+3.74)}}$
% & 94.07$_{\color{drop}{(-0.71)}}$
% & \underline{35.97}$_{\color{up}{(+13.40)}}$
% & 19.38$_{\color{up}{(+4.94)}}$
% & 10.86$_{\color{drop}{(-1.79)}}$
% \\

% & P-S + SC 	& 67.20$_{\color{drop}{(-0.80)}}$ 
% & \underline{85.68}$_{\color{up}{(+4.24)}}$
% & 94.37$_{\color{drop}{(-0.41)}}$
% & 35.57$_{\color{up}{(+13.00)}}$
% & \underline{19.83}$_{\color{up}{(+5.39)}}$
% & 11.01$_{\color{drop}{(-1.64)}}$
% \\

\midrule

\multicolumn{1}{c|}{ \multirow{7}{*}{ \textbf{MultiArith} } } 
% & FT SOTA 	& 60.5$^\gamma$ & \quad/ & \quad/ & \quad/ & \quad/ & \quad/ \\
% \multicolumn{1}{c}{/} & \multicolumn{1}{c}{/} & \multicolumn{1}{c}{/} & \multicolumn{1}{c}{/} & \multicolumn{1}{c}{/}  \\
% \cmidrule{2-8}

& CoT       & 97.67 
& \textbf{88.53} 
& \textbf{94.91} 
& 7.77 
& 7.47
& 5.51
\\

& CoT + SV 	& \textbf{98.33}$_{\color{up}{(+0.66)}}$ 
& 88.36$_{\color{drop}{(-0.17)}}$
& 94.38$_{\color{drop}{(-0.53)}}$
& \textbf{46.59}$_{\color{up}{(+38.82)}}$
& \textbf{24.56}$_{\color{up}{(+17.09)}}$
& \textbf{10.54}$_{\color{up}{(+5.03)}}$
\\

& CoT + SC 	& 98.17$_{\color{up}{(+0.50)}}$ 
& 88.42$_{\color{drop}{(-0.11)}}$
& 94.82$_{\color{drop}{(-0.09)}}$
& 10.22$_{\color{up}{(+2.45)}}$
& 9.29$_{\color{up}{(+1.82)}}$
& 5.33$_{\color{drop}{(-0.18)}}$
\\

\cmidrule{2-8}

% & ZSL CoT 	& 64.83 
% & 89.37 
% & 95.34 
% & 47.92 
% & 25.13 
% & 10.70
% \\

& ZS CoT        & 87.00     
& \underline{89.32} 
& 95.30  
& \underline{47.54} 
& 24.39 
& \underline{10.75} 
\\

& ZS CoT + SV   & \underline{97.00}$_{\color{up}{(+10.00)}}$ 
& 88.35$_{\color{drop}{(-0.97)}}$
& 94.38$_{\color{drop}{(-0.92)}}$
& 46.26$_{\color{drop}{(-1.28)}}$
& \underline{24.58}$_{\color{up}{(+0.19)}}$
& 10.54$_{\color{drop}{(-0.21)}}$
\\

& ZS CoT + SC   & \underline{97.00}$_{\color{up}{(+10.00)}}$ 
& 89.18$_{\color{drop}{(-0.14)}}$
& \underline{95.32}$_{\color{up}{(+0.02) \sim}}$
& 47.42$_{\color{drop}{(-0.12)}}$
& 23.83$_{\color{drop}{(-0.56)}}$
& 10.63$_{\color{drop}{(-0.12)}}$
\\

% & P-S 		& 76.17     
% & 86.31 
% & \underline{95.17}  
% & 22.89 
% & 15.34 
% & \underline{13.48} 
% \\

% & P-S + SV 	& 78.33$_{\color{up}{(+2.16)}}$ 
% & 86.67$_{\color{up}{(+0.36)}}$
% & 94.06$_{\color{drop}{(-1.11)}}$
% & \underline{40.45}$_{\color{up}{(+17.56)}}$
% & \underline{23.18}$_{\color{up}{(+7.84)}}$
% & 11.10$_{\color{drop}{(-2.38)}}$
% \\

% & P-S + SC 	& \underline{86.00}$_{\color{up}{(+9.83)}}$ 
% & \underline{87.36}$_{\color{up}{(+1.05)}}$
% & 94.44$_{\color{drop}{(-0.73)}}$
% & 38.83$_{\color{up}{(+15.94)}}$
% & 23.00$_{\color{up}{(+7.66)}}$
% & 11.34$_{\color{drop}{(-2.14)}}$
% \\

\midrule

\multicolumn{1}{c|}{ \multirow{7}{*}{ \textbf{MathQA} } } 

& CoT       & 52.83 
& \textbf{85.99}
& \textbf{95.31} 
& 49.57 
& 23.78 
& \textbf{7.64}  
\\

& CoT + SV  & \textbf{54.74}$_{\color{up}{(+1.91)}}$ 
& 85.93$_{\color{drop}{(-0.06)}}$
& 95.24$_{\color{drop}{(-0.07)}}$
& 51.39$_{\color{up}{(+1.82)}}$
& 24.61$_{\color{up}{(+0.83)}}$
& 7.18$_{\color{drop}{(-0.46)}}$
\\

& CoT + SC  & 54.47$_{\color{up}{(+1.64)}}$ 
& 85.93$_{\color{drop}{(-0.06)}}$
& 95.20$_{\color{drop}{(-0.11)}}$
& \textbf{51.73}$_{\color{up}{(+2.16)}}$
& \textbf{25.03}$_{\color{up}{(+1.25)}}$
& 7.15$_{\color{drop}{(-0.49)}}$
\\

% % \cdashline{3-8}[1pt/1pt]
\cmidrule{2-8}

% & ZSL CoT     & 63.68 
% & -
% & - 
% & - 
% & - 
% & - 
% \\

& ZS CoT        & 49.45 
& 85.20
& \underline{96.08} 
& 23.50 
& 13.76 
& 13.44  
\\

& ZS CoT + SV   & \underline{52.86}$_{\color{up}{(+3.41)}}$ 
& \underline{85.93}$_{\color{up}{(+0.73)}}$
& 95.24$_{\color{drop}{(-0.84)}}$
& \underline{51.40}$_{\color{up}{(+27.90)}}$
& \underline{24.63}$_{\color{up}{(+10.87)}}$
& 7.19$_{\color{drop}{(-6.25)}}$
\\

& ZS CoT + SC   & 49.51$_{\color{up}{(+0.06)}}$ 
& 85.22$_{\color{up}{(+0.02)} \sim}$
& 96.08$_{\color{drop}{(-0.00)} \sim}$
& 23.66$_{\color{up}{(+0.16)}}$
& 13.79$_{\color{up}{(+0.03)}}$
& \underline{13.48}$_{\color{up}{(+0.04)} \sim}$
\\

% & P-S       & 42.18    
% & - 
% & -  
% & - 
% & - 
% & - 
% \\

% & P-S + SV   & -$_{\color{drop}{(-0.01)}}$ 
% & -$_{\color{drop}{(-0.01)}}$
% & -$_{\color{drop}{(-0.01)}}$
% & -$_{\color{drop}{(-0.01)}}$
% & -$_{\color{drop}{(-0.01)}}$
% & -$_{\color{drop}{(-0.01)}}$
% \\

% & P-S + SC   & 42.55$_{\color{up}{(+0.37)}}$ 
% & -$_{\color{drop}{(-0.01)}}$
% & -$_{\color{drop}{(-0.01)}}$
% & -$_{\color{drop}{(-0.01)}}$
% & -$_{\color{drop}{(-0.01)}}$
% & -$_{\color{drop}{(-0.01)}}$
% \\

\midrule

\multicolumn{1}{c|}{ \multirow{7}{*}{ \textbf{CSQA} } } 
% \multicolumn{1}{c|}{ \multirow{7}{*}{ \textbf{Commonsense} } } 
% \multicolumn{1}{c|}{ \multirow{7}{*}{ \tabincell{c}{ \textbf{Commonsense} \\ (CSQA) \\} } } 
% & FT SOTA 	& \textbf{91.2}$^\eta$ & \quad/ & \quad/ & \quad/ & \quad/ & \quad/ \\
% \cmidrule{2-8}

& CoT       & 74.77 
& 81.40
& 92.57 
& \textbf{95.57} 
& \textbf{57.54} 
& 2.46 
\\

& CoT + SV 	& 74.04$_{\color{drop}{(-0.73)}}$ 
& 80.89$_{\color{drop}{(-0.51)}}$
& 92.10$_{\color{drop}{(-0.47)}}$
& 92.77$_{\color{drop}{(-2.80)}}$
& 56.05$_{\color{drop}{(-1.49)}}$
& \textbf{2.47}$_{\color{up}{(+0.01) \sim}}$
\\

& CoT + SC 	& \textbf{75.27}$_{\color{up}{(+0.50)}}$ 
& \textbf{81.50}$_{\color{up}{(+0.10)}}$
& \textbf{92.71}$_{\color{up}{(+0.14)}}$
& 95.04$_{\color{drop}{(-0.53)}}$
& 56.97$_{\color{drop}{(-0.57)}}$
& 2.43$_{\color{drop}{(-0.03)}}$
\\

\cmidrule{2-8}

% & ZSL CoT 	& 66.99 
% & 79.87
% & 95.26 
% & 25.62 
% & 29.43 
% & 9.87 
% \\

& ZS CoT        & 67.57     
& \underline{79.77} 
& \underline{95.26}  
& \underline{25.81} 
& 29.17 
& \underline{9.90} 
\\

& ZS CoT + SV   & 66.42$_{\color{drop}{(-1.15)}}$ 
& 79.06$_{\color{drop}{(-0.71)}}$
& 94.65$_{\color{drop}{(-0.61)}}$
& 25.36$_{\color{drop}{(-0.45)}}$
& 28.56$_{\color{drop}{(-0.61)}}$
& 9.06$_{\color{drop}{(-0.84)}}$
\\

& ZS CoT + SC   & \underline{71.58}$_{\color{up}{(+4.01)}}$ 
& 79.51$_{\color{drop}{(-0.26)}}$
& 95.21$_{\color{drop}{(-0.05)} \sim}$
& 25.08$_{\color{drop}{(-0.73)}}$
& \underline{29.69}$_{\color{up}{(+0.52)}}$
& 8.96$_{\color{drop}{(-0.94)}}$
\\

% & P-S 		& 64.21     
% & 76.05 
% & 93.74  
% & 18.92 
% & \underline{26.46} 
% & \underline{9.93} 
% \\
% & P-S + SV 	& 62.74$_{\color{drop}{(-1.47)}}$ 
% & 78.10$_{\color{up}{(+2.05)}}$
% & 94.44$_{\color{up}{(+0.70)}}$
% & \underline{22.50}$_{\color{up}{(+3.58)}}$
% & 26.09$_{\color{drop}{(-0.37)}}$
% & 9.77$_{\color{drop}{(-0.16)}}$
% \\

% & P-S + SC 	&  \underline{65.93}$_{\color{up}{(+1.72)}}$ 
% & \underline{78.13}$_{\color{up}{(+2.08)}}$
% & \underline{94.47}$_{\color{up}{(+0.73)}}$
% & 22.32$_{\color{up}{(+3.40)}}$
% & 26.11$_{\color{drop}{(-0.35)}}$
% & 9.62$_{\color{drop}{(-0.31)}}$  
% \\

\bottomrule
\end{tabular}
}

\vspace{-2mm}

\caption{
Comprehensive performance (\%) with different strategies on GPT-3.5 (\texttt{gpt-3.5-turbo}). 
% Results on MathQA are presented in Appendix~\ref{sec:appendix_result_mathqa}. 
% \textbf{FT SOTA}: Fine-tuning SOTA; % Previous 
% $\alpha$: \cite{arXiv2021_PreSOTA4GSM8K}, 
% $\beta$: \cite{EMNLP2022_PreSOTA4SVAMP}, 
% $\gamma$: \cite{EMNLP2015_PreSOTA4MultiArith},
% % $\delta$: , 
% $\eta$: \cite{IJCAI2022_PreSOTA4CSQA};
\textbf{CoT}: Few-shot CoT \cite{NeurIPS2022_CoT} with complex-prompting \cite{ICLR2023_ComplexCoT}; % \cite{NeurIPS2022_CoT} 
\textbf{ZS-CoT}: Zero-Shot CoT \cite{NeurIPS2022_ZeroShotCoT}; 
% \textbf{P-S}: \emph{Zero-shot} Plan-and-Solve following \citet{ACL2023_Plan-and-Solve}; 
\textbf{SV}: Self-Verification \cite{EMNLP2023-Findings_Self-Verification};
\textbf{SC}: Self-Consistency \cite{ICLR2023_Self-Consistency}. 
\textbf{Best few-shot results} are marked in \textbf{bold};  
\underline{best \emph{zero-shot} results} are \underline{underlined}. 
I/O: input/output. 
$\uparrow$: larger is better. 
% {\color{up}{\textasciitilde}}, {\color{drop}{\textasciitilde}}: comparable. 
{\color{up}{$\sim$}}, {\color{drop}{$\sim$}}: comparable. 
% $\sim$
\label{tab:exp_comprehensive}
} 

% \vspace{-1.3mm}
\vspace{-4mm}

\end{table*}

\section{Experiments}
\label{sec:experiment}

% \subsection{Datasets \& Models}
\subsection{Setup}

\textbf{Evaluation Metrics.} 
In this paper, we aim to conduct comprehensive evaluation on {\task}, thus we select some scores from \texttt{ROSCOE} \cite{ICLR2023_ROSCOE} as introduced in \S\ref{sec:ROSCOE}, which contains a suite of metrics allowing us to evaluate the quality of reasoning rationales, not limited to the correctness of final answers. 
% \citet{ICLR2023_ROSCOE} have proposed \texttt{ROSCOE}, a suite of metrics for {\task}. 
% The calculation of \texttt{ROSCOE} scores are introduced in Appendix~\ref{sec:appendix_roscoe}. 
% for space limits, 

\textbf{Datasets.} 
We evaluate on five benchmark datasets involving arithmetic and commonsense {\task}: 
\textbf{GSM8K} \cite{2021_GSM8K}, \textbf{SVAMP} \cite{NAACL2021_SVAMP}, \textbf{MultiArith} \cite{EMNLP2015_MultiArith}, \textbf{MathQA} \cite{NAACL2019_MathQA} and \textbf{CSQA} \cite{NAACL2019_CommonsenseQA}. 
% Due to space limits, we present MathQA experiments in Appendix~\ref{sec:appendix_result_mathqa}. 
% The size of each test set are presented in Table~\ref{tab:datasets}. 

% \input{tables/tab_datasets}

\textbf{Models.} 
% We compare different strategies on reasoning {\classA} and {\classB}. 
For reasoning \emph{\classA}, we leverage \textbf{Zero-shot CoT (ZS CoT)} \cite{NeurIPS2022_ZeroShotCoT} and \textbf{Few-shot CoT (CoT)} \cite{NeurIPS2022_CoT} with complexity-based prompting \cite{ICLR2023_ComplexCoT}. 
% , and present the experiments in Appendix~. , and \textbf{Plan-and-Solve} \cite{ACL2023_Plan-and-Solve} model 
For \emph{\classB}, we employ \textbf{Self-Verification (SV)} \cite{EMNLP2023-Findings_Self-Verification} and \textbf{Self-Consistency (SC)} \cite{ICLR2023_Self-Consistency} on multiple paths. 
SV is a step-level strategy, which verifies intermediate-step answers and returns the path containing the maximum number of correct step answers. 
SC is a path-level strategy, which conducts majority voting on final answers of all generated paths and selects the most consistent result. 
% highest frequency

\begin{figure*}[!t] % !htbp
  \centering
  \includegraphics[width=0.97\linewidth]{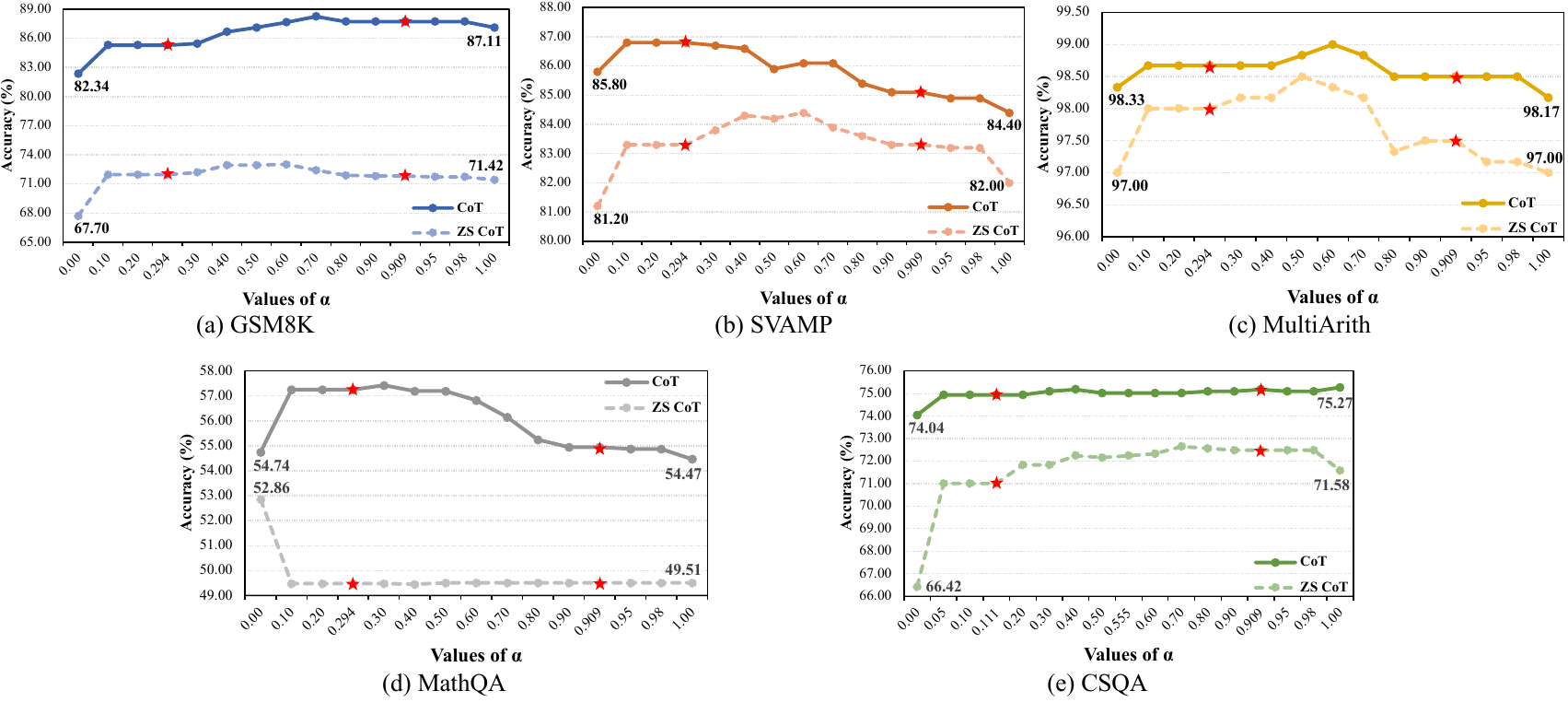}
  \vspace{-5mm}
  \caption{Accuracy under different integrated \emph{step-level} and \emph{path-level} {\classB} strategies, varying with the values of $\alpha$ defined in Eq~\eqref{eq:measure_score}. 
  Performance with two thresholds of $\frac{1}{ \frac{M(N-2)}{N} + 1 }$ and $\frac{1}{ \frac{1}{N} + 1}$ are marked as {\color{red}$\bigstar$}. 
  \label{fig:exp_joint} }
  \vspace{-5mm}
\end{figure*}

\textbf{Implementation.} 
We release the codes and generated results anonymously\footnote{\url{https://github.com/231sm/Eval_Multi-Step_Reasoning}.}. 
% anonymously\footnote{\url{https://anonymous.4open.science/r/Eval_Multi-Step_Reasoning-4E60}.}.
% on GitHub\footnote{\url{https://github.com/231sm/Eval_Multi-Step_Reasoning}.}. 
% anonymously, on the anonymous platform, https://anonymous.4open.science/r/Eval_Multi-Step_Reasoning-4E60
% For answer generation, we use GPT-3.5 (200B) with \texttt{gpt-3.5-turbo} engine as the backbone LLM. 
In this paper, the number of reasoning paths $N$ defined in Eq~\eqref{eq:measure_score} is 10, and number of intermediate steps $M$ is 3 on all datasets except for CSQA where $M$ is 10. 
% following \cite{EMNLP2023-Findings_Self-Verification}
% On CSQA dataset, $M$ is 10, as we conduct step-level {\classB} for 10 times similar to \cite{EMNLP2023-Findings_Self-Verification}. 
We utilize GPT-3.5 with \texttt{gpt-3.5-turbo} engine as the backbone LLM to generate reasoning paths (the model choice justification is elaborated in Appendix~\ref{sec:model_choice}), and the temperature is set to 0.7. 
We also leverage GPT-4 \cite{OpenAI_GPT4} with \texttt{gpt-4} engine to generate ground-truth rationales given the ground-truth answers for all datasets excluding GSM8K (which already contains them). 
For evaluation referring to \texttt{ROSCOE} \cite{ICLR2023_ROSCOE}, we respectively leverage \texttt{all-MiniLM-L6-v2}/\textit{SentenceTransformer}, and pretrained \texttt{gpt2-large} \cite{OpenAI2019_GPT2} to obtain token/sentence embedding and calculate perplexity defined in Eq~\eqref{eq:perplexity_path}. % \footnote{\url{}
All the reasoning paths for CoT and ZS CoT were generated during 8th to 23rd June 2023, and {\classB} on the generated reasoning paths was conducted during 12th October to 8th November 2023.

% \fontsize{10.5pt}{\baselineskip}\selectfont
\subsection{Analysis on Step-Level and Path-Level Answer Calibration Strategies} 
\label{sec:exp_strategy}

We respectively incorporate the effective step-level and path-level {\classB} strategies, Self-Verification (SV) and Self-Consistency (SC), into CoT-based models operating on multiple paths. We evaluate their performance using six evaluation metrics, with the results presented in Table~\ref{tab:exp_comprehensive}. 
% We equip CoT-based models with two {\classB} strategies on multiple paths: Self-Verification \& Self-Consistency, and evaluate their performance following six metrics. The results are presented in Table~\ref{tab:exp_comprehensive}. % comprehensive

% \input{tables/tab_exp_comprehensive}

Generally, \textbf{in terms of \emph{accuracy}, employing {\classB} is effective.} 
Seen from Table~\ref{tab:exp_comprehensive}, we find that models equipped with SV and SC obviously outperform vanilla methods, as both few-shot and zero-shot CoT employing SV/SC achieve significant accuracy improvements on almost all tasks. 
% , as both CoT + SC and ZS CoT + SC achieve significant accuracy improvements on almost all tasks. 
% models equipped with self-consistency obviously outperform, as both few-shot and zero-shot CoT show significant improvements across all tasks. 
% is obviously advantageous
% The self-consistency strategy conducts majority voting on the answers of  all generated reasoning paths, which improves calibration of LLM outputs, thus boosting accuracy. 
Notably, zero-shot CoT with SV and SC achieves much more significant outperformance of accuracy than few-shot settings on almost all tasks, demonstrating that \textbf{{\classB} is more effective in zero-shot settings}. 
As zero-shot CoT is relatively challenging due to the absence of task-specific in-context learning, {\classB} strategies essentially creating a feedback loop where the model assesses its own performance and adjusts accordingly, could help to mitigate biases and overfitting to specific patterns during inference, allowing the model to better generalize to new types of problems and datasets. 

Furthermore, \textbf{in terms of \emph{other metrics}, {\classB} can improve \emph{consistency} on arithmetic tasks but weakens \emph{faithfulness}, \emph{informativeness} and \emph{perplexity} on both arithmetic and commonsense tasks.} 
Observed from Table~\ref{tab:exp_comprehensive}, we find that SV and SC weaken the \emph{perplexity} score (16 out of 20 cases), suggesting that the rationale generated from multiple paths is more complex than that from a single path with CoT models. 
However, these two strategies improve \emph{consistency} scores on arithmetic tasks (10 out of 16 cases; 14 out of 16 cases), intuitively benefiting from multiple paths. 
% But these two {\classB} strategies are advantageous on arithmetic tasks regarding \emph{consistency} scores, 
As SV verifies answers for intermediate steps and SC considers answers for all paths, they naturally enhance consistency within steps and between input/output (I/O). 
Additionally, SV and SC worsen \emph{faithfulness} and \emph{informativeness} on almost all tasks (15 out of 20 cases for both). % enable vanilla models to achieve worse
The possible reason is that {\classB} on multiple paths focuses more on answer accuracy, while its increased complexity of its rationales tends to result in lower alignment and concordance between the source content and the output path. 
% The possible reason is that commonsense tasks require more semantic understanding, making majority voting or verification beneficial for generating a more relevant reasoning path. On the other hand, arithmetic tasks focus more on numerical calculation, where the alignment and concordance between source content and the output path are often sufficient with a single path. 
Generally, despite the benefits of employing SV and SC to CoT-based methods, the improvements are task-dependent and vary across different metrics. 
% Furthermore, while {\classB} strategies generally improve accuracy, they tend to worsen perplexity, faithfulness, and informativeness on most tasks. 

% There are trade-offs between the metrics, for instance, a method might improve accuracy but at the cost of increased perplexity. The choice of {\classB} strategies would depend on which metric is most valuable for a given application. 
% For example, if one needs the highest accuracy, they might opt for CoT + SV, but if one values consistency within steps, CoT + SC might be the preferred choice. The best model will vary depending on the specific requirements of the task at hand.

\iffalse
\bibliography{custom}
\bibliographystyle{acl_natbib}
\fi
% !TEX root = ./arr2024.tex

\begin{table*}[!t] % !htbp, !t
\centering
\small

% \resizebox{\linewidth}{!}{%
\begin{tabular}{c l | c c c c}

\toprule

\textbf{Engine} & \multicolumn{1}{l|}{ \textbf{Strategy} }
& \textbf{GSM8K} & \textbf{SVAMP} & \textbf{MultiArith} & \textbf{CSQA} \\

\midrule

\multirow{4}{*}{ \tabincell{c}{GPT-3 (175B) \\ \texttt{code-davinci-001}} } 
& CoT
& 13.84\;\; & 38.42\;\; & 45.85\;\; & 46.75\;\; \\
& CoT + SV
& 13.92${\color{up}{\uparrow}}$ & 38.96${\color{up}{\uparrow}}$ & 46.19${\color{up}{\uparrow}}$ & 47.68${\color{up}{\uparrow}}$ \\
& CoT + SC
& 23.40${\color{up}{\bm{\Uparrow}}}$ & 54.58${\color{up}{\bm{\Uparrow}}}$ & 79.82${\color{up}{\bm{\Uparrow}}}$ & 54.92${\color{up}{\bm{\Uparrow}}}$ \\
& CoT + SC + SV & 23.59${\color{up}{\bm{\Uparrow}}}$ & 54.68${\color{up}{\bm{\Uparrow}}}$ & 80.01${\color{up}{\bm{\Uparrow}}}$ & 55.09${\color{up}{\bm{\Uparrow}}}$
\\ 

\midrule

\multirow{4}{*}{ \tabincell{c}{Instruct-GPT (175B) \\ \texttt{code-davinci-002}} } 
& CoT
& 60.81\;\; & 75.87\;\; & 96.13\;\; & 77.42\;\; \\
& CoT + SV
& 65.14${\color{up}{\bm{\Uparrow}}}$ & 76.99${\color{up}{\uparrow}}$ & 99.15${\color{up}{\bm{\Uparrow}}}$ & 77.83${\color{up}{\uparrow}}$ \\
& CoT + SC
& 78.00${\color{up}{\bm{\Uparrow}}}$ & 86.77${\color{up}{\bm{\Uparrow}}}$ & 100.00${\color{up}{\bm{\Uparrow}}}$~~ & 81.43${\color{up}{\bm{\Uparrow}}}$ \\
& CoT + SC + SV & 78.32${\color{up}{\bm{\Uparrow}}}$ & 86.94${\color{up}{\bm{\Uparrow}}}$ & 100.00${\color{up}{\bm{\Uparrow}}}$~~ & 81.53${\color{up}{\bm{\Uparrow}}}$
% 99.80${\color{up}{\bm{\Uparrow}}}$
\\

\midrule

\multirow{4}{*}{ \tabincell{c}{GPT-3.5 \\ \texttt{gpt-3.5-turbo}} } 
& CoT
& 80.21\;\; & 78.20\;\; & 97.67\;\; & 74.77\;\; \\
& CoT + SV
& 82.34${\color{up}{\uparrow}}$ & 85.80${\color{up}{\bm{\Uparrow}}}$ & 98.33${\color{up}{\uparrow}}$ & 74.04${\color{drop}{\downarrow}}$ \\
& CoT + SC
& 87.11${\color{up}{\bm{\Uparrow}}}$ & 84.40${\color{up}{\bm{\Uparrow}}}$ & 98.17${\color{up}{\uparrow}}$ & 75.27${\color{up}{\uparrow}}$ \\
& CoT + SC + SV & 88.25${\color{up}{\bm{\Uparrow}}}$ & 86.80${\color{up}{\bm{\Uparrow}}}$ & 99.00${\color{up}{\bm{\uparrow}}}$ & 75.18${\color{up}{\bm{\uparrow}}}$
\\

\bottomrule

\end{tabular}
% }

\vspace{-2mm}

\caption{
Accuracy (\%) with different backbone engines. % performance 
${\color{up}{\uparrow}}$/${\color{up}{\bm{\Uparrow}}}$: slightly/significantly {\color{up}{better}};
${\color{drop}{\downarrow}}$: slightly {\color{drop}{worse}} than the baseline few-shot CoT. 
We refer to \citet{EMNLP2023-Findings_Self-Verification} for results with GPT-3 and Instruct-GPT engines. As \citet{EMNLP2023-Findings_Self-Verification} didn't test on MathQA dataset, we also exclude the results of MathQA here for fair comparisons. 
% Results of CoT+SV+SC with GPT-3.5 are the best performance in Figure~\ref{fig:exp_joint}. 
% ${\color{up}{\uparrow}}$/${\color{up}{\bm{\Uparrow}}}$, ${\color{drop}{\downarrow}}$/${\color{drop}{\bm{\Downarrow}}}$: performance is slightly/significantly {\color{up}{better}}, and slightly/significantly {\color{drop}{worse}} than the baseline CoT. 
% respectively denotes that 
\vspace{-6mm}
\label{tab:exp_effect_engines}
} 

\end{table*}

\subsection{Analysis on Unified Answer Calibration Strategies} 
\label{sec:exp_strategy_joint}

We then integrate step-level and path-level {\classB} strategies, varying $\alpha$ as defined in Eq~\eqref{eq:measure_score}. We present the accuracy of the unified strategies in Figure~\ref{fig:exp_joint}. 
As observed, accuracy peaks at a specific value of $\alpha$ between the two thresholds defined in Eq~\eqref{eq:measure_score_sv_sc_alpha_min_final} and \eqref{eq:measure_score_sc_sv_alpha_max_final} in almost all scenarios across all tasks (\ie, 8 out of 10 cases), demonstrating that \textbf{optimal model performance should balance both step-level and path-level {\classB} dominance}. 
Besides, we notice that for ``CoT on SVAMP task'' in Figure~\ref{fig:exp_joint}(b) and ``zero-shot CoT on MathQA task'' Figure~\ref{fig:exp_joint}(d), employing integrated {\classB} strategies reaches a peak with $\alpha$ not between the two thresholds, and the overall performance remains stably lower than the initial best accuracy with $\alpha = 0$ (\ie, SV). The possible reason may related to \emph{employing SV (\ie, $\alpha = 0$) presenting more significant advantages than SC (\ie, $\alpha = 1$) in the two scenarios}. 
Specifically, CoT on SVAMP respectively achieves accuracy of 85.80\% and 84.40\% when $\alpha$ values 0 (SV) and 1 (SC), with the difference larger than 1\%; 
Zero-shot CoT on MathQA employing SV and SC achieves accuracy of 52.86\% v.s. 49.51\%, where the difference is larger than 3\%. 
Except for these two distinctive scenarios, others in Figure~\ref{fig:exp_joint} obtain the optimal results by synthesizing step-level and path level {\classB} dominance. 

% relatively much better SV (\ie, $\alpha = 0$) performance than SC (\ie, $\alpha = 1$)
% possibly explaining why the overall performance of only these two integrated strategies remains stably lower than the initial best accuracy with $\alpha = 0$ (self-verification). 

% GSM8K (a): The accuracy of CoT is consistently higher than ZS CoT across all values of $\alpha$. The accuracy peaks at a specific value of $\alpha$ before declining.
% SVAMP (b): Both methods show a decline in accuracy as $\alpha$ increases, with ZS CoT exhibiting a more pronounced decrease.
% MultiArith (c): CoT maintains high accuracy across all $\alpha$ values, while ZS CoT sees a decrease after a certain point.
% MathQA (d): There's a peak in accuracy for CoT at a mid-range $\alpha$ value, after which it slightly declines. ZS CoT remains relatively stable but at a lower accuracy level.
% CSQA (e): CoT improves with increasing $\alpha$, reaching a plateau, while ZS CoT shows an initial increase followed by a slight decrease. 

In conclusion, the value of $\alpha$ plays a significant role in the performance of both few-shot and zero-shot CoT. Optimal ranges of $\alpha$ for each task are mostly between the two thresholds of step-level and path-level {\classB} dominance. The marked two thresholds represent boundaries for optimizing performance, which could guide further fine-tuning. 
Besides, the performance variance across datasets implies that the characteristics of each task, such as complexity, size, or the nature of the tasks. Models equipped with {\classB} strategies may require task-specific tuning to achieve the best performance. 

% % \noindent 
% In conclusion, the implications of these findings mainly include:
% (1) \emph{threshold significance}: The value of $\alpha$ plays a significant role in the performance of both few-shot and zero-shot CoT. Optimal ranges of $\alpha$ for each task are mostly between the two thresholds of step-level and path-level {\classB} dominance. The marked two thresholds represent boundaries for optimizing performance, which could guide further fine-tuning. 
% % calibrating the initial answers
% (2) \emph{self-verification superiority}: If the initial self-verification performance is significantly better than self-consistency, \eg, with accuracy difference larger than 1\%, the best performance may tend to achieve with step-level {\classB} strategies. 
% (3) \emph{task characteristics}: The performance variance across datasets implies that the characteristics of each task, such as complexity, size, or the nature of the tasks, impacting the effectiveness of the {\classB} strategies differently. Models may require task-specific tuning to achieve the best performance. 

\subsection{Effects of Backbone Models} 
\label{sec:exp_backbone}

% \textbf{Backbone LLM.} 
We compare accuracy on CoT-based {\classB} strategies with different LLM backbone engines, and present results in Table~\ref{tab:exp_effect_engines}. 

% \input{tables/tab_exp_effect_engines}

% the two {\classB} strategies
As observed from the results,  
for GPT-3 and Instruct-GPT, both self-verification (SV) and self-consistency (SC) provide consistent improvements; while on the larger GPT-3.5 model, their improvements are observably weaker, particularly for SV, with which accuracy even slightly drops on the CSQA task. 
The possible reason is that GPT-3.5 is more prone to making mistakes when verifying on intermediate-step answers for multiple paths. 
% We present the average step number of multiple paths compared with a single path in Appendix~\ref{sec:appendix_step_num}. We can observe that 
% Multiple paths generated by GPT-3.5 may constain more steps for each than GPT-3 and Instruct-GPT, 
% We suspect that the average number of steps for multiple paths generated by GPT-3.5 may be larger than that generated by GPT-3 and Instruct-GPT, making verification more fallible. 
Further, for integrated {\classB} strategies (SV+SC), the model's performance is close to the better one between SV and SC. % at intermediate values of $\alpha$ 
% the performance tendency with the backbone varying is consistent with the better one between self-verification and self-consistency. 
Generally, path-level {\classB} is more advantageous than step-level one, with relatively higher accuracy and lower computation cost. 
% If we employ self-verification (SV) and self-consistency (SC) to the backbone LLM engine of GPT-3.5, the outperformance is less significant or accuracy may even slightly drop. 
% We can infer that if the {\classA} for CoT with strong backbone LLM is sophisticated enough, the {\classB} may be simplified. We can directly implement \emph{path-level} {\classB} for multiple paths. 
Based on these observations, we can infer that \textbf{{\classB} strategies, especially path-level self-consistency, provide benefits in many cases, particularly on less powerful LLMs}. 

We further speculate, if the {\classA} for CoT with strong backbone LLM is sophisticated enough, the {\classB} may be simplified. We can directly conduct \emph{path-level} {\classB} for multiple paths. 
But these findings cannot indicate that step-level {\classB} is meaningless for stronger backbone LLMs. 
As seen from Table~\ref{tab:exp_comprehensive}, LLM equipped with step-level {\classB} is relatively beneficial to improve consistency scores. Besides, as mentioned in \citet{EMNLP2023-Findings_Self-Verification}, step-level {\classB} can provide explainable answers by verifying on intermediate-step answers, making results more reliable. 
% Particularly in the scenarios requiring high-reliability, step-level {\classB} is crucial. 
% and explainability, of answers

\begin{figure*}[!t] % [!htbp][!t]
  \centering
  \vspace{-5mm}
  \includegraphics[width=0.9\linewidth]{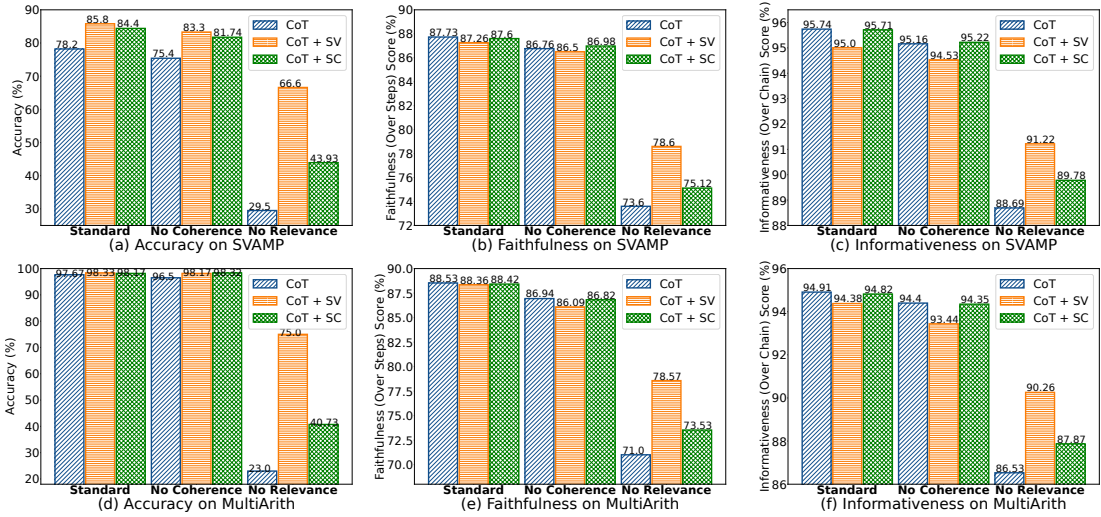}
  \vspace{-3mm}
  \caption{
  Performance (\%) of ``\textit{Accuracy}, \textit{Faithfulness (Over Steps)} and \textit{Informativeness (Over Path)}'' on SVAMP and MultiArith with different prompting on CoT models. 
  We didn't show full results of other tasks for space limits. 
  \label{fig:exp_effect_prompts} 
  }
  % \vspace{-8mm} % for local
  \vspace{-4mm} % for online
\end{figure*}

\subsection{Effects of Prompting} 
\label{sec:exp_prompting}

We further demonstrate the effects of prompting with few-shot demonstrations on {\classB}, evaluated on CoT models. 

We respectively input prompts of \emph{no coherence} and \emph{no relevance} for few-shot CoT referring to \citet{ACL2023_AnalyzeCoT} (examples are listed in Appendix~\ref{sec:appendix_cases_prompts}), and present performance on SVAMP and MultiArith in Figure~\ref{fig:exp_effect_prompts}. 
% Appendix~\ref{sec:appendix_cases_prompts}, Table~\ref{tab:examples_prompts}
% More results are shown in Appendix~\ref{sec:appendix_effect_prompts}. 
As seen, the deficiency of coherence and relevance in the prompting observably weaken the performance of all models, with no relevance having a more profound impact than no coherence. 
% prompts without coherence and relevance tend to weaken the performance, especially for no-relevance prompts. 
In addition, CoT+SV achieves comparable performance with CoT+SC when prompting is standard or not coherent. Further, CoT+SV tends to perform observably better than CoT+SC, when prompting with no relevance, indicating that step-level {\classB} strategy SV, is beneficial to maintain performance under adverse conditions. 
% In addition, CoT with self-verification outperforms when the prompting loses coherence or relevance.  
% indicating that step-level {\classB} strategy, specifically self-verification, can mitigate the degeneration caused by . 
This observation suggests \textbf{the robustness of step-level {\classB}}. It also highlights \textbf{the potential benefits of step-level {\classB} strategies to mitigate performance degeneration caused by poor prompting}. 
The possible reason is that step-level {\classB} strategies break down the task into subtasks, and these subtasks are simple enough so that less likely to be influenced by the low-quality prompts. 
% , and may be more effective in handling low-quality inputs. 
% This observation reveals that step-level {\classB}, especially self-verification, can mitigate the degeneration caused by poor prompting, and may be more effective in handling low-quality inputs. 
% Furthermore, 

\subsection{Analysis on Tasks} 
\label{sec:exp_tasks}
% % \noindent
% \textbf{Questions (Tasks).} 
% In Table~\ref{tab:exp_comprehensive}, we can observe that different tasks present differences in performance when equipping CoT with {\classB}. 
% Table~\ref{tab:exp_comprehensive} illustrates that the performance varies across different tasks when CoT is equipped with {\classB}. 
As seen from Table~\ref{tab:exp_comprehensive},\ref{tab:exp_effect_engines}, and Figure~\ref{fig:exp_joint}, generally, \textbf{SV and SC present more significant outperformance on arithmetic tasks than on the commonsense task (CSQA)}. 
Further, for CSQA, employing {\classB} tends to worsen the consistency scores, which is contrary to the trend observed in arithmetic tasks. 
% This could be because CSQA task focuses more on relevance and concordance between steps and paths, while arithmetic tasks prioritize coherence within steps and between I/O. 
The possible explanation lies in the characteristics of each task, such as complexity, size, or the nature of the tasks. 
In the CSQA task, correct intermediate steps may not always contribute to a coherent reasoning path due to potential irrelevance and redundancy. Specifically, even if we calibrate both intermediate step and path answers, there can be some correct commonsense statements while irrelevant to the question, resulting in worse consistency and perplexity. 
% , despite accuracy can improve. 
Conversely, in arithmetic tasks, correct intermediate answers almost guarantee a consistent reasoning path, as all intermediate answers are necessary and will contribute to a correct final answer. 
\section{Conclusion and Future Work}
\label{sec:con_fw}
% \vspace{-0.5mm}

In this paper, we dissect {\task} into {\classA} and {\classB}, and provide a unified view of {\classB} strategies through a comprehensive evaluation.
We find that path-level {\classB} is particularly potent in improving accuracy, while step-level {\classB} is more suitable for addressing issues related to low-quality prompting. The improvement is more pronounced in zero-shot scenarios and less significant on stronger backbone models. 
We also define step-level and path-level {\classB} dominance with two thresholds, and propose to integrate of the two types of strategies, which is promising to achieve optimal performance. 
% The best performance is often found between two thresholds, while predominant step-level {\classB} performance will make the best result deviate from this range. 
Our findings suggest that {\classB} is a versatile strategy that can be integrated into various models to bolster {\task} capabilities of LLMs. 
In the future, we aim to develop more sophisticated {\task} models, drawing on the insights and conclusions from this study. 
% and powerful 
% our focus is the development of more sophisticated {\task} models based on the conclusion of this paper. 

% In this paper, we has underscored the significant role that {\classB} strategies play in enhancing {\task} capabilities of LLMs. 

% Looking ahead, there are several promising directions for future research. One potential area of focus is the development of more sophisticated {\task} strategies that can better handle complex reasoning tasks and low-quality prompting. Another is the exploration of ways to integrate {\classB} with other techniques to further enhance the {\task} capabilities of LLMs.

% Furthermore, it would be interesting to investigate how {\classB} strategies can be adapted for use in other types of models and tasks. Finally, given the importance of accuracy, consistency, faithfulness, and informativeness in {\task}, future research should also explore ways to measure and improve these aspects in a more systematic and comprehensive manner.

% In summary, while our study has shed light on the potential of {\classB} strategies in enhancing {\task}, it has also highlighted the need for further research in this area. We hope that our findings will serve as a foundation for future studies aimed at advancing our understanding of {\task} and developing more effective strategies for improving its accuracy and reliability. 

\clearpage

\section*{Limitations}
The main limitation for this paper is that we didn't analyze more {\classB} strategies, such as step-/path-level methods on the single path, and varying the numbers of steps and paths in the unified {\classB} strategies. 
Besides, we can also employ {\classB} strategies to other {\classA} models, not limited to CoT-based methods. 
Further, we should also evaluate {\classB} strategies on more tasks to make the results more sufficient.

\section*{Broader Impact}

\textbf{Technical Novelty Emphasis.} 
We have conducted an empirical study of {\classB} and proposed a unified method to address that \emph{Step-Level / Path-Level Answer Calibration} for \emph{a Single or Multiple Paths} can be integrated together, with the two thresholds of \emph{Step-/Path-Level Dominant Answer Calibration} and a hyper-parameter $\alpha$. 
Our analysis has the potential to inspire further research and practical implications on unified answer calibration, such as \emph{``how the hyper-parameter $\alpha$ can be optimally chosen across different tasks, like iterative tuning''}. Our paper is based on an empirical study, and its main contributions are to unify multiple seemingly disparate types of approaches into a common framework, allowing us to investigate empirical questions to obtain more insights, \eg, 
% such as: 
\begin{enumerate}[(1)]
\item Employing {\classB} can enhance accuracy, with the improvement being more noticeable in zero-shot scenarios and less significant on stronger backbone models; 
\item The optimal performance of the unified {\classB} strategy typically achieved by synthesizing step-level and path level dominance;  
\item Path-level {\classB} is more beneficial in improving accuracy, and step-level {\classB} is more effective for mitigating low-quality prompting;   
\item Answer calibration can improve consistency on arithmetic tasks but weakens faithfulness, informativeness and perplexity on both arithmetic and commonsense tasks. 
\end{enumerate}

\section*{Acknowledgment}
We would like to express gratitude to the anonymous reviewers for their kind and helpful comments. 
This work was supported by the National Natural Science Foundation of China (No. 62206246), the Fundamental Research Funds for the Central Universities (226-2023-00138), Zhejiang Provincial Natural Science Foundation of China (No. LGG22F030011), Yongjiang Talent Introduction Programme (2021A-156-G), Tencent AI Lab Rhino-Bird Focused Research Program (RBFR2024003), Information Technology Center and State Key Lab of CAD\&CG, Zhejiang University, and NUS-NCS Joint Laboratory (A-0008542-00-00). 
% % We would like to express gratitude to the anonymous reviewers for their kind and helpful comments. 
% % This work was supported by the National Natural Science Foundation of China (No. 62206246), the Fundamental Research Funds for the Central Universities (226-2023-00138), Zhejiang Provincial Natural Science Foundation of China (No. LGG22F030011), Yongjiang Talent Introduction Programme (2021A-156-G), CCF-Baidu Open Fund, and Information Technology Center and State Key Lab of CAD\&CG, Zhejiang University, and NUS-NCS Joint Laboratory (A-0008542-00-00). 
% % This work was supported by the National Natural Science Foundation of China (No.62206246), Zhejiang Provincial Natural Science Foundation of China (No. LGG22F030011), Ningbo Natural Science Foundation (2021J190), Yongjiang Talent Introduction Programme (2021A-156-G), CCF Tencent Rhino-Bird Open Research Fund, and NUS-NCS Joint Laboratory (A-0008542-00-00). 
% % National Natural Science Foundation of China (No.62206246, 91846204 and U19B2027)
% % Information Technology Center and State Key Lab of CAD\&CG, Zhejiang University, 

% \balance
\interlinepenalty=10000
\bibliography{custom}
% \bibliographystyle{acl_natbib}

\iffalse
\bibliography{custom}
\bibliographystyle{acl_natbib}
\fi
% !TEX root = ./arr2024.tex

% \input{tables/tab_supp_cases_prompts}

% \newpage

\appendix

\section*{Appendices}
\label{sec:appendices}

\section{Terminology Clarification of Answer Calibration and Model Calibration}
\label{sec:appendix_clarification_calibration}

To avoid the confusion caused by the usage of the already-existing concept \emph{``calibration''}, we provide a terminology clarification. 
% Maybe the terminology of \emph{``calibration''} is likely to mislead readers. 
We emphasize that \textbf{\emph{``answer calibration''} defined in our paper differs from \emph{``model calibration''}} \cite{ICML2005_ModelCalibration,ICML2017_ModelCalibration,EMNLP2023_ModelCalibration,ICLR2024_ModelCalibration}.
\emph{``Answer Calibration''} refers to the post-processing methods applied to one or more reasoning path(s), to obtain a final answer. We categorize {\classB} methods as `step-level' if they break down the reasoning path(s) into their individual steps, and `path-level' otherwise. In most cases, \emph{``Answer Calibration''} is more akin to \emph{``Answer Correction''} \cite{TACL2024_Survey_Self-Correction}, involves correcting mistakes in the initial output. We did give a definition like this in the \emph{Abstract}, \emph{Introduction}, and we have already provided clear definitions of \emph{``Answer Calibration''} in \S\ref{sec:method}.

\section{Model Choice Justification}
\label{sec:model_choice}

The choice of GPT-3.5 was driven by its relevance and accessibility for our research objectives. Our research includes an empirical study of {\classB} and a proposal of a unified method, where the backbone LLM is pluggable. 
To facilitate reproducibility, we have already released the code and LLM-generated data anonymously\footnotemark[1] (provided at the bottom of Page 5 in \S\ref{sec:experiment}), aiming to enhance transparency to some extent and facilitate further research in this area. We remain committed to exploring more transparent models in future work.

\newpage
\section{Cases of Low-Quality Prompts}
\label{sec:appendix_cases_prompts}

We list some examples of prompts in Table~\ref{tab:examples_prompts}. 

\iffalse
\bibliography{custom}
\bibliographystyle{acl_natbib}
\fi
% !TEX root = ./arr2024.tex

\begin{table}[!htbp] % !htbp, !t
\centering
\small

\resizebox{\linewidth}{!}{

\begin{tabular}{l l}
\toprule

\textbf{Prompt Setting} & 
\tabincell{l}{ 
\emph{\textbf{Example Query (Arithmetic Reasoning)}} \\ 
\textit{Leah had 32 chocolates and her sister had 42.} \\
\textit{If they ate 35}, \\ 
\textit{how many pieces do they have left in total?} \\
} \\
% \midrule

% STD (Standard prompting) &  39  &  So the final answer is: Rudra Shah.  \\
 
\midrule

\textbf{Standard CoT} & 
\tabincell{l}{ 
Originally, \\ Leah had 32 chocolates and her sister had 42. \\ 
So in total they had 32 + 42 = 74. \\ 
After eating 35, \\ they had 74 - 35 = 39 pieces left in total. \\ 
The answer is 39. 
} \\ 
\midrule

\textbf{No Coherence} &  
\tabincell{l}{ 
After eating 32 + 42 = 74, \\ they had 32 pieces left in total. \\ 
Originally, Leah had 74 - 35 = 39 chocolates \\ and her sister had 35. \\ 
So in total they had 42. The answer is 39. 
} \\ 

\midrule

\textbf{No Relevance} & 
\tabincell{l}{ 
Patricia needs to donate 19 inches, \\ 
and wants her hair to be 31 inches long \\ after the donation. \\ 
Her hair is 29 inches long currently. \\ 
Her hair needs to be 19 + 31 = 50 \\ inc long when she cuts it. \\ 
So she needs to grow 50 - 29 = 21 more inches. \\
The answer is 21. 
} \\  

\bottomrule
\end{tabular}
}

\caption{
Examples of prompts (standard, no coherence and no relevance) in our experiments.
\label{tab:examples_prompts}
}
%\vspace{-1em}

\end{table}

\end{document}